\documentclass[runningheads]{llncs}

\PassOptionsToPackage{table}{xcolor}
\usepackage{eccv}

\usepackage{eccvabbrv}
\usepackage{graphicx}
\usepackage{booktabs}
\usepackage{multirow}
\usepackage{makecell}
\usepackage{arydshln}
\usepackage{tikz}
\usepackage{tcolorbox}
\usetikzlibrary{arrows.meta, positioning, shapes.geometric, fit, backgrounds, calc}

\usepackage{amsmath,amsfonts,bm}

\def\eqref#1{(\ref{#1})}

\def\1{\bm{1}}

\def\vh{{\bm{h}}}

\def\vw{{\bm{w}}}
\def\vx{{\bm{x}}}

\def\mA{{\bm{A}}}
\def\mB{{\bm{B}}}

\def\mW{{\bm{W}}}

\DeclareMathAlphabet{\mathsfit}{\encodingdefault}{\sfdefault}{m}{sl}
\SetMathAlphabet{\mathsfit}{bold}{\encodingdefault}{\sfdefault}{bx}{n}

\def\gC{{\mathcal{C}}}

\def\emA{{A}}

\newcommand{\R}{\mathbb{R}}

\definecolor{Gray}{gray}{0.93}

\newcommand{\bestcell}[1]{\textbf{#1}}
\newcommand{\htablegrayrow}[1]{\rowcolor{Gray}#1}

\definecolor{copied}{rgb}{0.858, 0.188, 0.478}

\newcommand{\iconfrozen}{\raisebox{-0.16ex}{\includegraphics[height=0.5em]{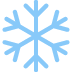}}}
\newcommand{\icontrainable}{\raisebox{-0.16ex}{\includegraphics[height=0.5em]{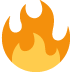}}}

\definecolor{fmcontrastive}{HTML}{2A78D6}
\definecolor{fmdino}{HTML}{1BAF7A}
\definecolor{fmmimft}{HTML}{EDA100}
\definecolor{fmmimssl}{HTML}{008300}
\definecolor{fmvlm}{HTML}{4A3AA7}

\newcommand{\mkbase}[2]{%
  \tikz[baseline=-0.55ex]{\draw[fill=#1,draw=white,line width=0.3pt] #2;}}
\newcommand{\mkcircle}[1]{\mkbase{#1}{(0,0) circle (0.48ex)}}
\newcommand{\mksquare}[1]{\mkbase{#1}{(-0.42ex,-0.42ex) rectangle (0.42ex,0.42ex)}}
\newcommand{\mkdiamond}[1]{\mkbase{#1}{(0,-0.55ex)--(0.5ex,0)--(0,0.55ex)--(-0.5ex,0)--cycle}}
\newcommand{\mktriangle}[1]{\mkbase{#1}{(-0.52ex,-0.36ex)--(0.52ex,-0.36ex)--(0,0.55ex)--cycle}}
\newcommand{\mkplus}[1]{\mkbase{#1}{%
  (-0.17ex,-0.5ex)--(0.17ex,-0.5ex)--(0.17ex,-0.17ex)--(0.5ex,-0.17ex)--
  (0.5ex,0.17ex)--(0.17ex,0.17ex)--(0.17ex,0.5ex)--(-0.17ex,0.5ex)--
  (-0.17ex,0.17ex)--(-0.5ex,0.17ex)--(-0.5ex,-0.17ex)--(-0.17ex,-0.17ex)--cycle}}

\newcommand{\mcontrastive}{\mkcircle{fmcontrastive}}   
\newcommand{\mvlm}{\mkplus{fmvlm}}                     
\newcommand{\mmimft}{\mkdiamond{fmmimft}}              
\newcommand{\mdino}{\mksquare{fmdino}}                 
\newcommand{\mmimssl}{\mktriangle{fmmimssl}}           

\usepackage{hyperref}
\usepackage{orcidlink}
\usepackage{placeins}
\usepackage{float} 
\usepackage{pifont} 
\newcommand{\cmark}{\ding{51}}
\newcommand{\xmark}{\ding{55}}

\usepackage[normalem]{ulem}
\colorlet{suggestnew}{green!50!black}
\colorlet{suggestold}{red!60!black}

\begin{document}

\title{
LoRA-based Adaptation Alone Is Not Enough: 
Understanding the Limits of Foundation Models for Face Presentation Attack Detection
}
\titlerunning{LoRA Adaptation Alone Is Not Enough for Face PAD}

\author{Peter Lorenz\inst{1} \and
Anjith George\inst{1} \and
Sébastien Marcel\inst{1,2}}

\authorrunning{P.~Lorenz et al.}

\institute{
Idiap Research Institute, Martigny, Switzerland
\and
University of Lausanne (UNIL), Lausanne, Switzerland
}

\maketitle

\begin{abstract}
Face presentation attack detection (PAD) aims to reliably detect a wide range of presentation attacks. While PAD methods achieve strong performance within individual datasets, their performance degrades under cross-dataset evaluation.
Variations in sensors or lighting conditions can reduce the effectiveness of detectors from near-perfect to nearly random.
Foundation models (FMs) have emerged as a promising alternative because typical PAD datasets, such as the MCIO benchmarks (MSU-MFSD, CASIA-FASD, Replay-Attack, and OULU-NPU), are small relative to the scale used for web-based pretraining.
However, existing PAD systems primarily focus on CLIP-based foundation models, while overlooking other FMs with different architectures and training procedures.
This study addresses this question by systematically evaluating 32 FMs.
Zero-shot prompting achieves performance near chance across model families and scales.
The vision encoders, when low-rank-adapted (LoRA) with fewer than 1\% trainable weights, achieve below 2\% intra-dataset ACER in most cases, while cross-dataset ACER is substantially higher.
LoRA primarily refines the decision boundary within a dataset, suggesting that pretrained representations and the adaptation dataset play a larger role in cross-dataset generalization than the evaluated lightweight adaptation strategy.
\href{https://paper.pages.idiap.ch/lora-pad}{paper.pages.idiap.ch/lora-pad}.
\end{abstract}

\section{Introduction}
\label{sec:intro}
Face recognition has advanced rapidly with deep learning~\cite{deng2019arcface,boutros2022elasticface}, yet it remains vulnerable to presentation attacks: a printed photograph, a replayed video, or a similar spoof can induce a recognition system to grant access to an impostor~\cite{OtroshiShahreza_IEEE-TIFS_2025,zhang2020celeba,DBLP:journals/pr/FangDKK22,pasmino2023flickr}, or to reject a genuine user.
Presentation attack detection (PAD) screens out spoofed presentations from genuine ones before verification~\cite{DBLP:conf/cvpr/LiuJ018,DBLP:journals/pami/YuWQLLZ21,DBLP:journals/tbbis/YuLSXZ21,Fang_2022_WACV}.
Detectors trained and tested on a single dataset achieve strong accuracy, but their performance drops sharply when the sensor, illumination, or attack type changes at test time~\cite{DBLP:conf/cvpr/ShaoLLY19,DBLP:conf/aaai/ShaoLY20,DBLP:conf/cvpr/JiaZSC20,DBLP:journals/tbbis/WangWDG22,DBLP:journals/tcsv/YanZH22,fang2024face}.
This cross-dataset gap, rather than intra-dataset accuracy, is the central open and long-standing problem.

Vision encoders and vision--language models (VLMs) pretrained on web-scale image or image--text data~\cite{radford2021learning,kirillov2023segment,oquab2023dinov2,chen2024internvl} encode substantially richer visual priors than the limited labelled data available in any PAD dataset~\cite{deng2009imagenet,fang2023synthaspoof,fang2024face}, and recent work adapts them to PAD through parameter-efficient tuning such as LoRA~\cite{gonzalez2025foundation,feng2026benchmarking}, multimodal architectures~\cite{srivatsan2023flip,zhang2025interpretable,lin2025instructflip}, and face-specific self-supervision~\cite{wang2025fsfm} (\cref{sec:related}).
Whether such adaptation closes the cross-dataset gap remains unclear, as few encoders have been compared under a common parameter-efficient protocol~\cite{ozgur2025foundpad,gonzalez2025foundation,feng2026benchmarking}.

\begin{figure}[tbp]
  \centering
  \includegraphics[width=\linewidth]{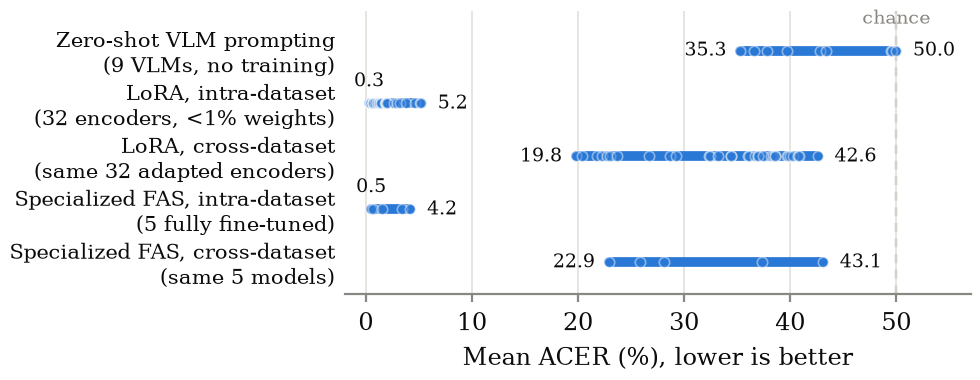}
  \caption{
        Mean ACER range across models (dots: models; bars: min--max).
        Zero-shot VLMs perform near chance (Q1,~\Cref{tab:zeroshot-intra-dataset-acer}), while LoRA achieves 0.3--5.2\% intra-dataset ACER (Q2,~\Cref{tab:lora-intra-dataset-acer}) but 19.8--42.6\% cross-dataset ACER across 32 encoders (Q3,~\Cref{tab:lora-cross-dataset-summary}), indicating that LoRA
        alone does not overcome dataset bias.
  }
  \label{fig:adaptation-ladder}
\end{figure}

This benchmark includes 32 pretrained vision encoders, comprising vision-only models and the vision towers extracted from VLMs, each adapted with the same LoRA recipe on the four datasets, together with nine complete VLMs prompted zero-shot on the same splits and metric.
We focus on LoRA as a widely used parameter-efficient method due to its simplicity and effectiveness~\cite{miao2025taso}.
Three specialist detectors serve as reference points from before the foundation-model era: DeepPixBiS~\cite{george2019deep}, a classical CNN, FSFM-FAS~\cite{wang2025fsfm}, a face-pretrained ViT, FoundPAD~\cite{ozgur2025foundpad}, a LoRA-based approach, and FLIP~\cite{srivatsan2023flip}.
The goal is to characterize how representations behave across model families rather than to optimize dataset-specific accuracy.

Our contributions, framed by a single question: \emph{is LoRA adaptation sufficient for robust PAD?}, are:
\begin{itemize}
\item \textit{A comprehensive and systematic evaluation of foundation models for face PAD,}
covering 32 vision encoders, 9 VLMs, specialist PAD baselines, and common datasets under a unified experimental protocol. 
\item \textit{A bias free generalization score $G$: }
We propose a metric $G$ that jointly captures intra- and cross-dataset skill as the geometric mean below chance,
which could be used as a future metric. 
\item \textit{Unveiling the limitation of LoRA adaptation for PAD,}
analyzing compute, feature-space analysis, and pretraining-scale comparisons besides accuracy.
\end{itemize}

\section{Related Work}
\label{sec:related}
Face presentation attack detection (PAD) has evolved from developing task-specific representations using limited PAD datasets to leveraging large-scale pretrained models, thereby enhancing robustness under unseen conditions.

\textit{Learning representations from PAD data.}
Early PAD methods developed representations directly from PAD datasets using supervised objectives and appearance cues such as texture, color, and frequency information~\cite{DBLP:conf/cvpr/LiuJ018}. Subsequent approaches enhanced discrimination through pixel-wise supervision~\cite{george2019deep,DBLP:journals/tbbis/YuLSXZ21}, frequency-domain and multi-scale learning~\cite{Fang_2022_WACV}, and patch-based formulations~\cite{DBLP:conf/cvpr/WangLYL22}. 
While these methods are effective within the training distribution, their performance often declines in cross-dataset evaluations because they capture dataset-specific properties, such as illumination and attack artifacts, rather than learning transferable PAD representations~\cite{DBLP:journals/tbbis/WangWDG22,DBLP:journals/tcsv/YanZH22,fang2024face}.

To enhance generalization, subsequent research has explored self-supervised learning, domain adaptation, and domain generalization. Self-supervised methods reduce reliance on annotations by constructing supervisory signals from facial data, whereas domain adaptation leverages target-domain information during training~\cite{DBLP:journals/tifs/LiLCWHK18,DBLP:conf/icb/WangHSC19}. In contrast, domain generalization seeks to learn invariant representations from multiple source domains~\cite{DBLP:conf/cvpr/ShaoLLY19,DBLP:conf/aaai/ShaoLY20,DBLP:conf/cvpr/JiaZSC20,DBLP:conf/aaai/ChenYSDTLHJ21}. The use of auxiliary datasets and synthetic attack generation further increases data diversity~\cite{zhang2020celeba,fang2023synthaspoof}. Despite these efforts, multi-source evaluations consistently reveal substantial performance degradation on unseen datasets~\cite{DBLP:conf/cvpr/ShaoLLY19,feng2026benchmarking}, indicating that PAD-specific data alone may not yield sufficiently transferable representations.

\textit{Adapting large-scale pretrained representations for PAD.}
Recent research has examined whether large-scale pretrained representations offer greater transferability than models trained exclusively on PAD data. Vision foundation models pretrained through supervised, self-supervised, or multimodal objectives provide broad visual priors that can be adapted to PAD using various strategies~\cite{radford2021learning,oquab2023dinov2}. 
Feng et al.~\cite{feng2026benchmarking} evaluated pretrained architectures under multi-source PAD protocols and demonstrated that self-supervised Vision Transformers generalize more effectively than supervised CNNs. 
Additional improvements have been achieved through architectural and training modifications, such as register tokens~\cite{darcet2024registers}, PAD-specific augmentations, and patch supervision~\cite{cai2024fasaug,watanabe2022pda}.

In addition to full fine-tuning, parameter-efficient adaptation methods seek to retain pretrained knowledge by updating only a small subset of parameters. FoundPAD~\cite{gonzalez2025foundation} adapts CLIP using LoRA~\cite{hu2022lora}, while vision-language methods such as FLIP~\cite{srivatsan2023flip}, I-FAS~\cite{zhang2025interpretable}, and InstructFLIP~\cite{lin2025instructflip} incorporate language supervision.
FaceCoT~\cite{Zhang_2026_CVPR} explores multimodal large language models (MLLMs) by introducing a vision–language reasoning dataset and a chain-of-thought (CoT) enhanced learning strategy to improve spoof-detection robustness and interpretability.
Other strategies include prompt learning~\cite{liu2024cfpl}, frequency-aware modeling~\cite{cao2025towards}, domain-modality alignment~\cite{yang2025dadm}, and optimal-transport adaptation~\cite{li2025optimal}. 
FSFM-FAS~\cite{wang2025fsfm} investigates face-specific masked modeling pretraining, while few-shot multimodal studies~\cite{DBLP:conf/wacv/KomatyOGM25} indicate that semantic priors alone are insufficient for zero-shot PAD.

However, current foundation-model-based PAD studies assess only a limited range of backbone architectures, which complicates the separation of the effects of pretrained representations from those of adaptation strategies. 
FoundPAD and FLIP incorporate CLIP variants, while Feng et al.~\cite{feng2026benchmarking} compare a broader set of architectures but rely on full fine-tuning. 
As a result, it remains uncertain which pretraining paradigms, such as contrastive, self-supervised, or masked-image modeling, yield the most transferable representations for cross-dataset PAD and whether parameter-efficient adaptation can fully leverage these representations.

\section{Approach}
\label{sec:approach}

\subsection{Low-rank Adaptation (LoRA) of the Vision Encoders}
\label{sec:lora-adaptation}

Vision encoders are evaluated using LoRA and a linear head, i.e., a fully connected layer (FC).
Each cropped face image~$\vx$ is processed by a pretrained vision backbone with fixed weights during both training and testing, except for low-rank adapters applied to the attention projections.
The backbone produces a global image embedding $\vh \in \R^{d}$, with the extraction method determined by the specific backbone family.
Extraction methods include CLS token, post-layer-norm CLS, global average pooling, projected CLIP image embedding, or mean-pooled patch tokens.
Details are provided in the FC column of~\cref{tab:lora-intra-dataset-acer}.
The FC layer maps~$\vh$ to a scalar logit, $z = \vw^{\top} \vh + b$.
Training minimizes binary cross-entropy loss on~$z$, using bonafide labels $y{=}1$ and attack labels~$y{=}0$.
At inference, the bonafide probability is computed as $p=\sigma(z)$.

Following prior work~\cite{gonzalez2025foundation}, LoRA~\cite{hu2022lora} reparameterizes each frozen projection $\mW_0$ with a trainable low-rank update, $\mW' = \mW_0 + \frac{\alpha}{r} \mB\mA$, where rank $r{=}8$, scaling $\alpha{=}16$, and dropout is set to $0.05$. The matrix $\mB$ is initialized to zero, ensuring adaptation begins from the pretrained behavior.
Adapters are attached to the query and value projections ($\mW_Q$, $\mW_V$) of every attention block (\texttt{pwconv} layers for ConvNeXt; see the LoRA-site column in \cref{tab:lora-intra-dataset-acer}) and are trained jointly with the linear head.
This approach maintains the trainable parameter footprint below 1\% of the backbone weights while enabling the representation, rather than only the readout, to specialize for PAD.



\subsection{Vision Encoders Family and Data Scale}
\label{sec:encoderfamily}

\Cref{tab:encoder-taxonomy} presents the 32 models, organized by family and scale, along with their pretraining provenance and data scale.
Each backbone receives the same LoRA adaptation as described in \cref{sec:lora-adaptation}.
ViT-style backbones, including CLIP, DINOv2/v3, BEiT, SigLIP, ViT-MAE/MSN, and InternViT, utilize the final CLS token.
Vision towers extracted from multimodal large language models (Qwen-VL, LLaVA, DeepSeek-VL2) apply mean pooling over patch tokens, while ConvNeXt-V2 applies global average pooling.

The models span five broad pretraining categories (\cref{tab:encoder-pretraining-data}): \mcontrastive vision-language contrastive learning, which learns a shared embedding space by aligning paired images and text (CLIP~\cite{radford2021learning}, SigLIP~\cite{zhai2023sigmoid}, EVA-CLIP~\cite{sun2024evaclip18b}); \mdino self-distillation, which learns visual representations by enforcing consistency between different views or teacher--student networks (DINOv2~\cite{oquab2023dinov2}, its register-token variant DINOv2-R~\cite{darcet2024registers}, DINOv3~\cite{simeoni2025dinov3}); \mmimft masked-image transformer methods, which learn from partially observed images through masked prediction or representation matching (BEiT~\cite{bao2021beit}, ViT-MAE~\cite{he2022masked}, ViT-MSN~\cite{assran2022masked}); \mmimssl masked-image self-supervised convolutional models, represented by ConvNeXt-V2~\cite{woo2023convnext}, which combines a ConvNet architecture with fully convolutional masked autoencoding; and \mvlm vision towers from multimodal large language models, which encode images into visual representations that are subsequently integrated with a language model (InternViT~\cite{chen2024internvl}, Qwen-VL~\cite{wang2024qwen2vl,bai2025qwen25vl}, LLaVA-NeXT~\cite{liu2024llavanext}, LLaVA-OneVision~\cite{li2024llavaonevision}, DeepSeek-VL2~\cite{wu2024deepseekvl2}).
These vision towers from the VLMs maintain parameter counts between 0.3B and 0.8B.
For DeepSeek-VL2, the vision encoder is also extracted. 
DeepSeek‑VL2's vision encoder is a hybrid backbone that combines SigLIP‑L and SAM‑B~\cite{DBLP:conf/iccv/KirillovMRMRGXW23}.
SigLIP‑L, trained with an image–text contrastive objective, captures high-level semantic concepts that align with language, and operates on a resolution of 384 pixels. 
SAM-B, trained for segmentation or mask prediction, captures low-level structure, edges, textures, and fine spatial details, and operates on high-resolution images (1024$\times$1024 pixels). 
Its outputs are mapped by a vision‑language adaptor into a sequence of visual tokens that are then fed into the MoE large language model (LLM).
DeepSeek-VL2‑small and VL2‑7B use the same SigLIP vision encoder.
They differ only in the MoE language backbone size (2.8B vs 4.5B active parameters).

Three representative models are tracked throughout the experiments: \textbf{CLIP} (contrastive), \textbf{DINOv2-R} (self-supervised), and \textbf{LLaVA-NeXT} (vision-language model tower). 
More precisely, the multimodal LLaVA-NeXT v1.6 Vicuna with 7B parameters includes the pre-trained CLIP‑ViT‑Large (ViT‑L/14) with a 336$\times$336 pixel input resolution. In contrast, the other CLIP variants use a 224$\times$224 pixel input resolution.
LLaVA-Next system trains in two stages: 
1) Language-image alignment: 
The 2‑layer ReLU MLP projects CLIP image features into the LLM's token space.
2) Visual instruction tuning:
The full model (LLM, CLIP vision encoder, projector) is trained together but with different learning rates.
CLIP is subsequently fine‑tuned end‑to‑end with the LLM using a next‑token prediction loss on multimodal instruction data, with a conservative learning rate to keep the LLM stable.

\begin{table*}[tbp]
  \centering
  \caption{Pretraining provenance of the vision backbones in
  Tab.~\ref{tab:lora-intra-dataset-acer}. 
  Variants listed together (e.g.\ base/large) share
  the same pretraining recipe and data. \emph{Data scale} is the pretraining corpus size as
  reported by the underlying publication. 
  VLM token budgets cover the full multimodal pretraining,
  of which the vision tower sees only the visual share. 
  Params reports the frozen vision  backbone size in billions (B), per listed variant. 
  {\small
  SSL: self-supervised learning,
  MIM: masked image modeling, NTP: next-token prediction, FT: fine-tuning.
  }
  }
  \label{tab:encoder-pretraining-data}
  \label{tab:encoder-taxonomy}
  \footnotesize
  \resizebox{\linewidth}{!}{%
  \begin{tabular}{lrlll}
    \toprule
    Model & \makecell[c]{Params\\(B)} & Descripton & Pretraining data & Data scale \\
    \midrule
    CLIP ViT-B/32 or 16, ViT-L/14~\cite{radford2021learning}   & 0.09, 0.30 & Contrastive image--text & WIT-400M~\cite{radford2021learning} & 400M pairs \\
    CLIP ViT-H/14~\cite{schuhmann2022laion}                    & 0.63 & Contrastive image--text & LAION-2B~\cite{schuhmann2022laion} (English subset of LAION-5B) & 2.3B pairs, 32B seen \\
    BEiT-base, -large~\cite{bao2021beit}                       & 0.09, 0.30 & MIM (dVAE tokens), supervised FT & ImageNet-22K $\rightarrow$ ImageNet-1K~\cite{deng2009imagenet} & 14M images \\
    ConvNeXt-V2-base, -huge~\cite{woo2023convnext}             & 0.09, 0.66 & FCMAE (SSL), supervised FT & ImageNet-1K~\cite{deng2009imagenet} & 1.28M images \\
    ViT-MAE-base, -huge~\cite{he2022masked}                    & 0.09, 0.63 & Masked autoencoding (SSL) & ImageNet-1K~\cite{deng2009imagenet} & 1.28M images \\
    ViT-MSN-base, -large~\cite{assran2022masked}               & 0.09, 0.30 & Masked siamese networks (SSL) & ImageNet-1K~\cite{deng2009imagenet} & 1.28M images \\
    DINOv2-base, -giant~\cite{oquab2023dinov2}                 & 0.09, 1.14 & DINO+iBOT SSL (base distilled from ViT-g) & LVD-142M~\cite{oquab2023dinov2} (curated web images) & 142M images \\
    DINOv2-R-base, -giant~\cite{darcet2024registers}           & 0.09, 1.14 & DINOv2 SSL + register tokens & LVD-142M~\cite{oquab2023dinov2} & 142M images \\
    DINOv3-base, -huge+~\cite{simeoni2025dinov3}               & 0.09, 0.84 & SSL + Gram anchoring (distilled from ViT-7B) & LVD-1689M~\cite{simeoni2025dinov3} (curated Instagram/web pool) & 1.69B images \\
    SigLIP-base, -large~\cite{zhai2023sigmoid}                 & 0.09, 0.32 & Sigmoid contrastive image--text & WebLI~\cite{chen2023pali} (English subset) & $\sim$10B images \\
    EVA-CLIP-8B~\cite{sun2024evaclip18b}                       & 8.22 & Contrastive image--text (EVA init) & Merged-2B~\cite{sun2023eva} (LAION-2B~\cite{schuhmann2022laion} $\cup$ COYO-700M~\cite{byeon2022coyo}) & $\sim$2B pairs \\
    InternViT-300M~\cite{chen2024internvl2}                    & 0.30 & Cosine distillation from InternViT-6B, NTP & InternVL~2.x multimodal corpus~\cite{chen2024internvl2} & --- \\
    InternViT-6B~\cite{chen2024internvl2}                      & 5.54 & CLIP contrastive, then incremental NTP & LAION-en/-multi~\cite{schuhmann2022laion}, COYO~\cite{byeon2022coyo}, Wukong~\cite{gu2022wukong}; InternVL corpus~\cite{chen2024internvl} & $\sim$5B pairs \\
    Qwen2-VL-72B-ViT~\cite{wang2024qwen2vl}                    & 0.70 & DFN ViT init, multimodal NTP & DFN-5B~\cite{fang2024dfn}; proprietary mix (OCR, interleaved, VQA, video) & 1.4T tokens \\
    Qwen2.5-VL-3B-, 7B-ViT~\cite{bai2025qwen25vl}              & 0.67, 0.68 & ViT from scratch (CLIP), multimodal NTP & DataComp~\cite{gadre2023datacomp} + in-house; proprietary multimodal mix & 4.1T tokens \\
    Qwen3-VL-2B-, 32B-ViT                                      & 0.41, 0.60 & Multimodal NTP (no tech report) & Not disclosed & --- \\
    LLaVA-NeXT-7B-ViT~\cite{liu2024llavanext,radford2021learning} & 0.32 & CLIP ViT-L/14-336 tower, lightly tuned in SFT & WIT-400M~\cite{radford2021learning}; $\sim$1.3M instruction samples & 400M pairs \\
    LLaVA-OneVision-7B-ViT~\cite{li2024llavaonevision}         & 0.41 & SigLIP-SO400M tower, unfrozen in later stages & WebLI~\cite{chen2023pali}; $\sim$9.4M OneVision samples & --- \\
    DeepSeek-VL2-small-, 7B-ViT~\cite{wu2024deepseekvl2}       & 0.65, 0.78 & SigLIP-SO400M init, VL pretraining & WebLI~\cite{chen2023pali} + (WiT~\cite{srinivasan2021wit}, WikiHow~\cite{koupaee2018wikihow}, OBELICS~\cite{laurencon2023obelics}, in-house) & 800B tokens \\
    \bottomrule
  \end{tabular}}
\end{table*}

\subsection{Specialist PAD Baselines}
\label{sec:baselines}

To contextualize the LoRA results, we compare them with four end-to-end specialist PAD models spanning classical CNN supervision and SSL-pretrained ViT fine-tuning: DeepPixBiS~\cite{george2019deep}, a compact DNN with pixel-wise supervision.
FSFM-FAS~\cite{wang2025fsfm}, a ViT-B/16 fine-tuned after face-specific SSL pretraining on the dataset VGGFace2~\cite{cao2018vggface2,huang2022adaptive}.
FLIP~\cite{srivatsan2023flip} introduces three variants that differ in how much of the pretrained vision–language model is adapted: 
FLIP‑V fine‑tunes only the vision encoder.
FLIP‑IT fine-tunes both the image and text encoders.
FLIP‑MCL further fine‑tunes the non‑linear projection layer and both encoders, using language guidance to improve cross‑dataset generalization.
All are fully fine-tuned on each dataset under the same protocol splits as our adapted backbones.
FoundPAD~\cite{ozgur2025foundpad} investigates LoRA-fine-tuned transformers, mainly limited to ViT-B/16 and ViT-L/14 on the leave-one-out protocol.

\section{Experiments}
\label{sec:experiments}


\subsection{Datasets and Protocol}

This study evaluates only the single-source protocol, in which a model is trained on one dataset and tested either on the same dataset (intra-dataset) or on a different dataset (cross-dataset). 
Multi-source protocols, including leave-one-out and limited source domains, are not considered in this analysis.
The four MCIO datasets~\cite{wen2015face, costa2016replay, patel2016secure, boulkenafet2017OULU}: MSU-MFSD\footnote{M: Protocol \texttt{protocol} (all folds)
\href{https://files.pythonhosted.org/packages/45/18/3e2b324468d3cdc290ac08b73f8972fa2aa87bc1071605057acfe3dd9900/bob.db.msu_mfsd_mod-2.2.9.zip}{\texttt{bob.db.msu\_mfsd\_mod-2.2.9.zip}}} (\textbf{M}), CASIA-FASD\footnote{C: Protocol \texttt{grandtest}
\href{https://www.idiap.ch/software/bob/data/bob/bob.pad.face/pad-face-casia-fasd-0b07ea45.tar.gz}{\texttt{pad-face-casia-fasd.tar.gz}}} (\textbf{C}), Replay-Attack\footnote{I: Protocol \texttt{grandtest}
\href{https://www.idiap.ch/software/bob/data/bob/bob.pad.face/pad-face-replay-attack-aca6b46f.tar.gz}{\texttt{pad-face-replay-attack.tar.gz}}} (\textbf{I}), and OULU-NPU\footnote{O: Protocol \texttt{Protocol\_1}
\href{https://www.idiap.ch/software/bob/data/bob/bob.pad.face/pad-face-oulunpu-7bfb90c9.tar.gz}{\texttt{pad-face-oulunpu.tar.gz}}} (\textbf{O}) span laboratory and mobile capture with print, replay, warped- and cut-photo, and multi-sensor attacks under varying illumination, giving the cross-dataset shift typical of PAD benchmarking~\cite{DBLP:conf/icmcs/LiuCDLZX22,fang2024face}.
After face detection and cropping, the train/dev/test splits contain
3{,}196/3{,}037/2{,}398 (M), 3{,}600/1{,}200/7{,}200 (C),
7{,}199/7{,}200/9{,}600 (I), and 24{,}000/17{,}999/12{,}000 (O) frames,
with attacks outnumbering bonafide frames by roughly $3$--$5\times$ in every split.

\textbf{Pre-processing.}
The MTCNN\footnote{MTCNN weights \href{https://github.com/timesler/facenet-pytorch}{github.com/timesler/facenet-pytorch}.}~\cite{MTCNN} face detection preprocessing pipeline is employed to crop faces to a size of $224{\times}224$ pixels from the video frames.
A maximum of 20 frames are sampled at evenly spaced intervals throughout the video.
Each backbone model is paired with its corresponding image processor to perform resizing, normalization, and family-specific preprocessing.
The most common input size is $224{\times}224$ pixels.
However, this parameter varies across models. For example, InternViT uses $448{\times}448$ inputs, whereas SigLIP-Large uses $256{\times}256$ inputs.

\textbf{Models.}
In the LoRA experiments (\cref{tab:lora-intra-dataset-acer}), the low-rank adapters and the linear head are trained.
The five specialist PAD models are fully fine-tuned (\cref{sec:baselines}) on our datasets.

\textbf{Implementation details.}
We optimize the LoRA adapters and the linear head with Adam (binary cross-entropy, class-balanced mini-batches, batch size 32, $\mathrm{lr}{=}10^{-4}$ with cosine decay, early stopping on validation loss).
The Bob toolkit~\cite{bob2012, bob2017} computes the EER on the development set and ACER metrics on the test set.

\subsection{Evaluation Metrics}

The \emph{Average Classification Error Rate} (ACER) is reported in accordance with ISO/IEC~30107-3~\cite{ISO301073} and established PAD protocols~\cite{DBLP:conf/icmcs/LiuCDLZX22,Fang_2022_WACV,fang2024face}.
ACER is defined as the mean of attack and bonafide error rates, calculated at a threshold $\tau$ determined on the development split using the Equal Error Rate (EER) criterion. This approach, standard in PAD evaluation, ensures balanced error rates and is applied without modification to the test split.
\begin{equation}
  \mathrm{ACER}(\tau)
  = \tfrac{1}{2}\bigl(\mathrm{APCER}(\tau) + \mathrm{BPCER}(\tau)\bigr).
  \label{eq:acer}
\end{equation}
All ACER values are reported as percentages, where lower values indicate better performance.

Let $\gC=\{\mathrm{M},\mathrm{C},\mathrm{I},\mathrm{O}\}$ denote the datasets and $\emA_{s,t}$ the test ACER (Eq.~\eqref{eq:acer}) of a model trained on dataset~$s$ and evaluated on dataset~$t$.
From the resulting transfer matrix $\mA \in \R^{4 \times 4}$ we report three unweighted means:
\begin{equation}
  \mathrm{ACER}_{\mathrm{ID}}
  = \tfrac{1}{4} \textstyle\sum_{s} \emA_{s,s},
  \quad
  \mathrm{ACER}_{\mathrm{CD}}
  = \tfrac{1}{12} \textstyle\sum_{s \neq t} \emA_{s,t},
  \quad
  \mathrm{ACER}_{\mathrm{ALL}}
  = \tfrac{1}{16} \textstyle\sum_{s,t} \emA_{s,t},
  \label{eq:acer-aggregates}
\end{equation}
over the four intra-dataset (ID) pairs~\cite{boulkenafet2017OULU,george2019deep}, the twelve cross-dataset (CD) transfer pairs~\cite{DBLP:conf/icb/PereiraAMM13,DBLP:conf/icb/WangHSC19,DBLP:conf/cvpr/YuZWQ0LZZ20}, and all sixteen test evaluations (ID and CD combined). 
Each $\emA_{s,t}$ is computed on the test split of dataset~$t$.

To summarize joint intra- and cross-dataset metrics in a single number, we report the \emph{generalization score}
\begin{equation}
  G \;=\; \frac{\sqrt{\,(C-\mathrm{ACER}_{\mathrm{ID}})_+\;(C-\mathrm{ACER}_{\mathrm{CD}})_+\,}}{C},
  \qquad C=50,\;\; (x)_+ = \max(0,x),
  \label{eq:gscore}
\end{equation}
the geometric mean of ID and CD \emph{skill} (percentage points below the $C{=}50\%$ chance ACER), normalized to $[0,1]$: $G{=}1$.
Intuitively, $G$ captures how far a model is from chance in both settings simultaneously. 
Since the product of two positive numbers is zero if either factor is zero, a model performing at chance on ID will necessarily have $G=0$, regardless of its  CD ACER.
The score rises only when skill is present on both fronts. 
\Cref{tab:lora-cross-dataset-summary} reports ID, CD, and $G$ per model.

\subsection{Zero-Shot VLM Prompting Performance}
\label{sec:zeroshot}

\Cref{tab:zeroshot-intra-dataset-acer} presents results for prompting the VLM setting. 
No PAD-specific training is performed in this evaluation.
\begin{tcolorbox}[
    colback=gray!10,
    colframe=gray!10,
    boxrule=0pt,
    arc=8pt,
    left=10pt,
    right=10pt,
    top=8pt,
    bottom=8pt,
    fontupper=\ttfamily\small,
    boxsep=5pt,
    title={\textbf{Prompt}},
    coltitle=black,
    titlerule=2pt,
    before title={\vspace{-1pt}},
    after title={\vspace{-10pt}}
]
"Analyze the following image and determine whether it is bonafide or a presentation attack.
Must provide an authenticity score along with 'bonafide' or 'attacked' on a scale from 0 to 1, 
in the format: score: xx,  where xx is a number between 0 and 1. No extra text."
\end{tcolorbox}
In this configuration, full vision-language models (VLMs) are loaded using quantized NF4 precision and employ \texttt{bf16} or \texttt{fp16} computation.
Each test image is evaluated with the fixed prompt (see above).

\textit{Are VLMs capable of solving PAD without task-specific training?}
Among the nine evaluated VLMs, which span four model families and range from 2 billion to 32 billion total parameters (\cref{tab:zeroshot-intra-dataset-acer}), the average ACER remains between 35\% and 50\%, which is close to chance performance.
\textit{The answer is no}: language-aligned semantic reasoning does not substitute for task-aligned adaptation of the vision tower, even though it is the very same tower. 

\begin{table}[tpb]
  \centering
  \caption{
  Zero-shot VLM PAD: intra-dataset test set ACER$\downarrow$ (\%). 
  Prec.\ reports weight storage and compute dtype (e.g.\ NF4/\texttt{bf16}: 4-bit NF4 weights with \texttt{bfloat16} matmuls; \texttt{bf16}/\texttt{fp16}: full-precision loading). 
  Avg.\ is the unweighted mean, and Std.\ is the standard deviation across the datasets.
  }
  \label{tab:zeroshot-intra-dataset-acer}
  \footnotesize
  \resizebox{0.7\linewidth}{!}{%
  \begin{tabular}{lcccccccc}
    \toprule
    Model & Prec. & M & C & I & O & Avg. & Std. \\
    \midrule
    DeepSeek-VL2-small & NF4/fp16 & 43.8 & 22.4 & 46.9 & 45.7 & 39.7 & 10.1 \\
    DeepSeek-VL2 & NF4/fp16 & 50.9 & 38.1 & 49.3 & 32.9 & 42.8 & 7.5 \\
    Qwen2.5-VL-7B & NF4/bf16 & 49.9 & 47.9 & 50.1 & 50.0 & 49.5 & 0.9 \\
    Qwen2.5-VL-3B & NF4/bf16 & 50.0 & 50.0 & 50.0 & 50.0 & 50.0 & \bestcell{0.0} \\
    Qwen3-VL-2B & NF4/bf16 & 52.3 & 28.8 & \bestcell{42.0} & 50.3 & 43.4 & 9.3 \\
    Qwen3-VL-32B & NF4/bf16 & \bestcell{37.0} & 19.8 & 48.4 & 46.1 & 37.8 & 11.2 \\
    LLaVA-NeXT-7B & NF4/bf16 & 49.6 & 48.9 & 50.0 & 49.8 & 49.6 & 0.4 \\
    LLaVA-OneVision-7B & NF4/bf16 & 38.8 & 30.8 & 44.8 & \bestcell{32.1} & 36.6 & 5.6 \\
    \htablegrayrow{InternVL3-14B & NF4/bf16 & 38.4 & \bestcell{14.4} & 48.8 & 39.6 & \bestcell{35.3} & 12.7} \\
    \bottomrule
  \end{tabular}
  }
\end{table}

\subsection{Intra-dataset Performance under LoRA}
\label{sec:lora-intra-dataset}

\Cref{tab:lora-intra-dataset-acer} extends this recipe to 32 vision backbones (\cref{tab:encoder-taxonomy}), from CLIP and DINOv3 to the vision towers of multimodal LLMs.
The intra-dataset gains are large and consistent: CLIP ViT-B/32 reaches 0.3\% mean ACER, LLaVA-NeXT-7B-ViT 0.4\%, InternViT-6B 0.7\%, and most adapted backbones remain below 2\% across the four datasets while updating fewer than 1\% of the foundation model parameters.
{These results indicate that parameter-efficient adaptation is sufficient for strong intra-dataset PAD performance}, although, as discussed in \cref{sec:backbone-analysis}, the choice of backbone remains an important factor.

\begin{table}[tbp]
  \centering
  \caption{
  ID test ACER$\downarrow$ (\%) with LoRA fine-tuning: each cell trains and tests on the same dataset (threshold tuned on the \texttt{dev} split, EER criterion). 
  Features: adapter site (LoRA) and backbone feature fed to the fully connected head (FC). 
  Parameters: frozen backbone size (\iconfrozen, $\times10^{9}$) and trainable LoRA/FC parameter counts (\icontrainable, $\times10^{3}$). 
  Avg./Std.: unweighted mean and standard deviation across the datasets.
  \textbf{Bold}: lowest ACER per column.
  }
  \label{tab:lora-intra-dataset-acer}
  \footnotesize
  \resizebox{\linewidth}{!}{%
  \begin{tabular}{lllrrccccccc}
    \toprule
    \multirow{2}{*}{Model} & \multicolumn{2}{c}{Features} & \multicolumn{3}{c}{Parameters} & \multicolumn{6}{c}{Test Datasets} \\
    \cmidrule(lr){2-3} \cmidrule(lr){4-6} \cmidrule(lr){7-12}
    & \tiny{LoRA} & \tiny{FC} & {\tiny \iconfrozen $\times 10^9$} &  {\tiny\icontrainable(}\tiny{LoRA},& \tiny{FC)$\times 10^3$} & {M} & {C} & {I} & {O} & {Avg.} & {Std.} \\
    \midrule
    \multicolumn{12}{c}{Model Zoo - LoRA Finetuned} \\
    \midrule
               ViT-MSN-large     & Q,V & CLS & 0.30 & 786 & 1.0 & 4.2 & 3.2 & \textbf{0.0} & 13.2 & 5.1 & 4.9 \\
               Qwen3-VL-32B-ViT  & Q,V & Mean pool & 0.60 & 995 & 5.1 & 9.2 & 0.5 & 5.3 & 4.9 & 5.0 & 3.1 \\
               InternViT-300M    & Q,V & CLS & 0.30 & 786 & 1.0 & 3.5 & 0.9 & 1.3 & 11.3 & 4.2 & 4.2 \\
               ViT-MAE-base      & Q,V & CLS & 0.09 & 294  & 0.8 & 1.4 & 1.6 & 0.4 & 10.9 & 3.6 & 4.3 \\
               Qwen2.5-VL-3B-ViT & Q,V & Mean pool & 0.67 & 1310 & 2.0 & 3.8 & 0.5 & 1.2 & 6.9 & 3.1 & 2.5 \\
               ViT-MSN-base      & Q,V & CLS & 0.09 & 294 & 0.8 & 0.8 & 1.0 & 0.1 & 9.7 & 2.9 & 3.9 \\
               ViT-MAE-huge      & Q,V & CLS & 0.63 & 1310 & 1.3 & 2.8 & 0.6 & 0.1 & 5.4 & 2.2 & 2.1 \\
               Qwen2-VL-72B-ViT  & Q,V & Mean pool & 0.70 & 1310 & 8.2 & 1.8 & 1.2 & 4.4 & 0.8 & 2.1 & 1.4 \\
               DeepSeek-VL2-7B-ViT & Q,V & Mean pool & 0.78 & 995 & 2.6 & 4.3 & 0.1 & 3.4 & 0.5 & 2.1 & 1.8 \\
               DINOv2-R-base     & Q,V & CLS & 0.09 & 294 & 0.8 & 0.6 & 2.1 & 0.6 & 4.7 & 2.0 & 1.7 \\
               BEiT-large        & Q,V & LN CLS & 0.30 & 786 & 1.0 & 0.6 & 0.6 & 0.5 & 6.1 & 1.9 & 2.4 \\
               Qwen3-VL-2B-ViT   & Q,V & Mean pool & 0.41 & 786 & 2.0 & 2.5 & 0.3 & 2.9 & 1.7 & 1.9 & 1.0 \\
               DINOv2-giant      & Q,V & CLS & 1.14 & 1966 & 1.5 & 2.3 & 0.5 & 0.5 & 3.9 & 1.8 & 1.4 \\
               Qwen2.5-VL-7B-ViT & Q,V & Mean pool & 0.68 & 1310 & 3.6 & 2.9 & 0.2 & 2.3 & 0.3 & 1.4 & 1.2 \\
               SigLIP-large      & Q,V & LN CLS & 0.32 & 786 & 1.0 & 2.7 & 0.7 & 0.9 & 1.2 & 1.4 & 0.8 \\
               DINOv2-R-giant    & Q,V & CLS & 1.14 & 1966 & 1.5 & 2.9 & 0.2 & 1.7 & 0.8 & 1.4 & 1.0 \\
               DINOv2-base       & Q,V & CLS & 0.09 & 294 & 0.8 & 1.5 & 1.6 & 0.3 & 2.1 & 1.4 & 0.6 \\
               SigLIP-base       & Q,V & LN CLS & 0.09 & 294 & 0.8 & 0.9 & 0.7 & 2.0 & 1.2 & 1.2 & 0.5 \\
               LLaVA-OneVision-7B-ViT & Q,V & Mean pool & 0.41 & 958.5 & 3.6 & 1.8 & 0.4 & 0.7 & 1.8 & 1.2 & 0.6 \\
               BEiT-base         & Q,V & LN CLS & 0.09 & 294 & 0.8 & 0.3 & 0.6 & 0.4 & 3.2 & 1.1 & 1.2 \\
               ConvNeXt-V2-base  & pwconv & GAP+LN & 0.09 & 1443 & 1.0 & 0.4 & 0.8 & 1.4 & 1.7 & 1.1 & 0.5 \\
               ConvNeXt-V2-huge  & pwconv & GAP+LN & 0.66 & 3970 & 2.8 & 1.8 & 0.5 & 0.3 & 0.9 & 0.9 & 0.6 \\
               DeepSeek-VL2-small-ViT & Q,V & Mean pool & 0.65 & 995 & 2.0 & 1.7 & 0.1 & \textbf{0.0} & 1.1 & 0.7 & 0.7 \\
               EVA-CLIP-8B       & Q,V & Img emb. & 8.22 & 5505 & 1.3 & 0.6 & 0.5 & 0.8 & 0.9 & 0.7 & 0.2 \\
               CLIP ViT-H/14     & Q,V & LN CLS & 0.63 & 1310 & 1.3 & 0.3 & \textbf{0.0} & 2.3 & 0.2 & 0.7 & 0.9 \\
               DINOv3-base       & Q,V & CLS & 0.09 & 294 & 0.8 & 0.6 & 0.5 & 1.2 & 0.5 & 0.7 & 0.3 \\
               InternViT-6B      & Q,V & CLS & 5.54 & 4608 & 3.2 & 0.6 & 0.1 & 1.5 & 0.4 & 0.7 & 0.5 \\
               DINOv3-huge+      & Q,V & CLS & 0.84 & 1310 & 1.3 & \textbf{0.0} & 0.0 & 0.7 & 1.0 & 0.4 & 0.4 \\
               CLIP ViT-B/16     & Q,V & LN CLS & 0.09 & 294 & 0.8 & 0.4 & 1.1 & 0.2 & \textbf{0.0} & 0.4 & 0.4 \\
               LLaVA-NeXT-7B-ViT & Q,V & Mean pool & 0.32 & 786 & 4.1 & 1.2 & \textbf{0.0} & 0.2 & 0.0 & 0.4 & 0.5 \\
               CLIP ViT-L/14     & Q,V & LN CLS & 0.30 & 786 & 1.0 & \textbf{0.0} & 0.1 & \textbf{0.0} & 1.4 & 0.4 & 0.6 \\
\htablegrayrow{CLIP ViT-B/32     & Q,V & LN CLS & 0.09 & 294 & 0.8 & 0.8 & 0.2 & 0.2 & \textbf{0.0} & \textbf{0.3} & 0.3} \\
    \midrule
    \multicolumn{12}{c}{Baseline -- PAD Specialist Models other than LoRA} \\
    \midrule
               FSFM-FAS     & --- & --- & --- & \multicolumn{2}{c}{86}   & 2.8 & 1.6 & 0.2 & 12.0 & 4.2 & 4.6 \\
               DeepPixBiS   & --- & --- & --- & \multicolumn{2}{c}{3}    & 7.9 & 2.0 & \bestcell{0.0} & 3.8 & 3.4 & 2.9 \\
               FLIP-V       & --- & --- & --- & \multicolumn{2}{c}{86}   & \bestcell{0.0} & \bestcell{0.4} & 0.8 & 4.7 & 1.5 & 1.9 \\
               FLIP-MCL     & --- & --- & --- & \multicolumn{2}{c}{169}  & 1.6 & 0.7 & 0.3 & 0.1 & 0.7 & 0.6 \\
\htablegrayrow{FLIP-IT      & --- & --- & --- & \multicolumn{2}{c}{149}  & 0.4 & 0.8 & 0.8 & \bestcell{0.1} & \bestcell{0.5} & \bestcell{0.3}} \\
    \bottomrule
  \end{tabular}
  }
\end{table}

\subsection{Which Backbones Benefit most from LoRA?}
\label{sec:backbone-analysis}

\begin{figure}[tbp]
  \centering
  \includegraphics[width=\linewidth]{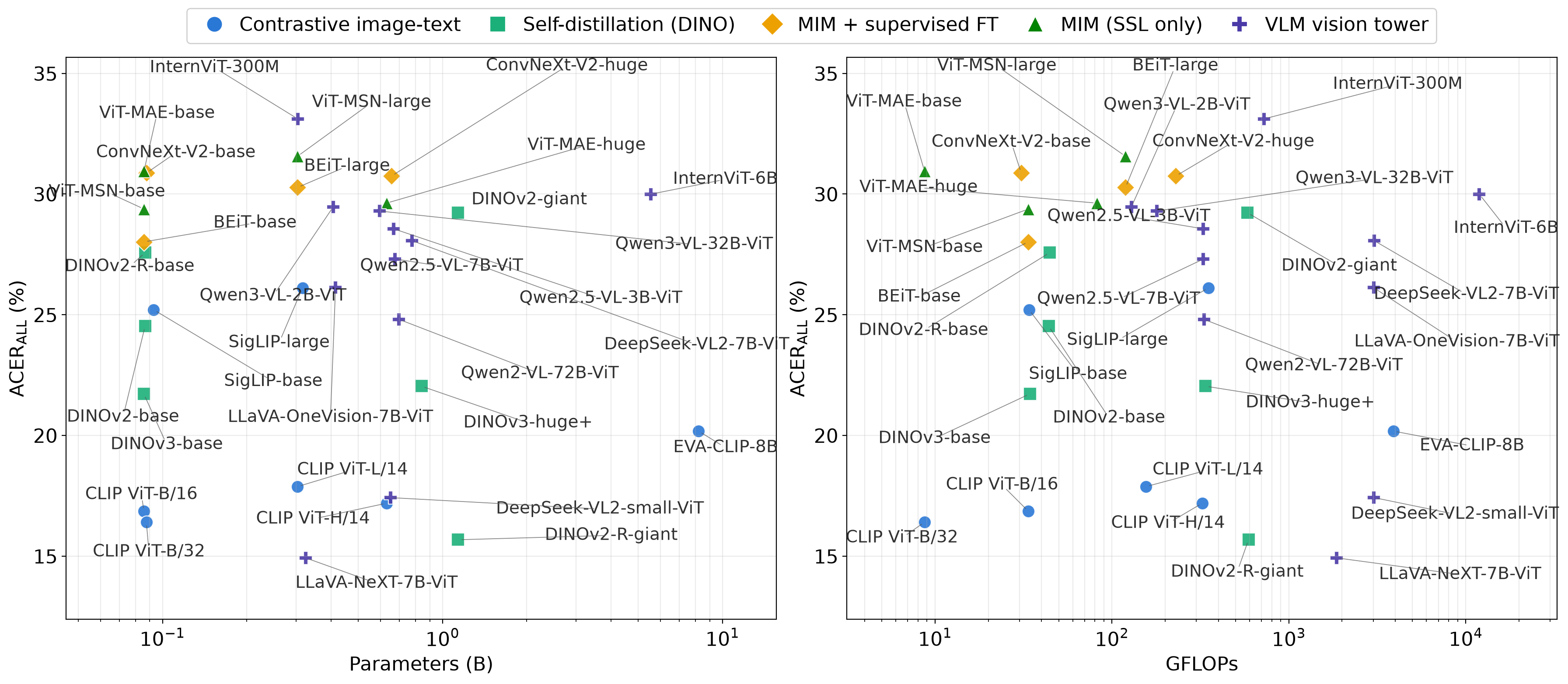}
  \caption{%
    ACER$_\mathrm{ALL}\downarrow$ (\%) vs. compute$\downarrow$ (GFLOPs).
    Left: frozen vision backbone parameters (billions).
    Right: forward-pass GFLOPs at batch size~1, where CLIP VIT-B/32 yields the best trade-off between accuracy and compute.
    Note that some jitter was added to the x-axis of the datapoints such that they do not overlap.
    Legend explained in \cref{sec:encoderfamily}.
  }
  \label{fig:compute-efficiency}
\end{figure}

\Cref{fig:compute-efficiency} relates post-LoRA ACER to backbone size and forward-pass cost, addressing the second half of Q2.
The pretraining objective matters more than parameter count: CLIP ViT-B/32, with a 90M-parameter backbone, reaches 0.3\% mean intra-dataset ACER, whereas the 0.60B-parameter vision tower extracted from the Qwen3-VL-32B VLM averages 5.0\% (\cref{tab:lora-intra-dataset-acer}).
Contrastive vision--language backbones (CLIP, EVA-CLIP, SigLIP) and self-distillation models (DINOv3) respond most reliably to LoRA.
The vision towers extracted from multimodal LLMs show mixed results across scales.
Gains also vary by dataset: the weakest backbones (ViT-MAE, ViT-MSN, InternViT-300M) remain the weakest on OULU-NPU, the dataset with the strongest sensor diversity.
Backbone choice and pretraining objective therefore dominate LoRA outcomes. 
Scale alone does not.

\subsection{Cross-Dataset Transfer under LoRA}
\label{sec:lora-cross}

\begin{table}[tbp]
  \centering
  \caption{
      ACER (\%) over the datasets
      (\cref{eq:acer-aggregates}): 
      {ID} (four intra-dataset pairs),
      {CD} (twelve transfer pairs), 
      All (sixteen pairs; ID + CD), and generalization score
      $G$ (higher is better).
      The geometric mean of intra-dataset and cross-dataset skill relative to the $50\%$ chance ACER.
      FLIP (IT/MCL/V), FSFM-FAS, and DeepPixBiS are the fully fine-tuned FAS-specialist baselines (\cref{sec:baselines}).
      \textbf{Bold}: best per column (lowest ACER, highest $G$).
  }
  \label{tab:lora-cross-dataset-summary}
  \scriptsize
  \setlength{\tabcolsep}{2pt}
  \resizebox{0.77\linewidth}{!}{%
  \begin{tabular}{@{}lrrrr@{}}
    \toprule
    Model & All & {ID} & {CD} & $G\uparrow$ \\
    \midrule
    \htablegrayrow{LLaVA-NeXT-7B-ViT & \bestcell{14.9} & 0.4 & \bestcell{19.8} & \bestcell{0.77}} \\
    DINOv2-R-giant & 15.7 & 1.4 & 20.4 & 0.76 \\
    CLIP ViT-B/32 & 16.4 & \bestcell{0.3} & 21.8 & 0.75 \\
    CLIP ViT-B/16 (FoundPAD) & 16.9 & 0.5 & 22.3 & 0.74 \\
    CLIP ViT-H/14 & 17.2 & 0.7 & 22.7 & 0.73 \\
    DeepSeek-VL2-small-ViT & 17.4 & 0.7 & 23.0 & 0.73 \\
    CLIP ViT-L/14 (FoundPAD) & 17.9 & 0.4 & 23.7 & 0.72 \\
    EVA-CLIP-8B & 20.2 & 0.7 & 26.7 & 0.68 \\
    DINOv3-base & 21.7 & 0.7 & 28.7 & 0.65 \\
    DINOv3-huge+ & 22.0 & 0.4 & 29.2 & 0.64 \\
    DINOv2-base & 24.5 & 1.4 & 32.3 & 0.59 \\
    Qwen2-VL-72B-ViT & 24.8 & 2.0 & 32.4 & 0.58 \\
    SigLIP-base & 25.2 & 1.2 & 33.2 & 0.57 \\
    SigLIP-large & 26.1 & 1.4 & 34.4 & 0.55 \\
    LLaVA-OneVision-7B-ViT & 26.1 & 1.2 & 34.4 & 0.55 \\
    Qwen2.5-VL-7B-ViT & 27.3 & 1.2 & 36.0 & 0.52 \\
    DINOv2-R-base & 27.6 & 1.9 & 36.1 & 0.52 \\
    BEiT-base & 28.0 & 1.1 & 37.0 & 0.50 \\
    DeepSeek-VL2-7B-ViT & 28.1 & 2.0 & 36.7 & 0.50 \\
    Qwen2.5-VL-3B-ViT & 28.6 & 3.2 & 37.0 & 0.49 \\
    DINOv2-giant & 29.2 & 1.8 & 38.4 & 0.47 \\
    Qwen3-VL-32B-ViT & 29.3 & 5.1 & 37.4 & 0.48 \\
    ViT-MSN-base & 29.4 & 2.9 & 38.2 & 0.47 \\
    Qwen3-VL-2B-ViT & 29.5 & 1.9 & 38.6 & 0.47 \\
    ViT-MAE-huge & 29.6 & 2.7 & 38.6 & 0.46 \\
    InternViT-6B & 30.0 & 0.7 & 39.8 & 0.45 \\
    BEiT-large & 30.3 & 1.9 & 39.7 & 0.44 \\
    ConvNeXt-V2-huge & 30.7 & 0.9 & 40.7 & 0.43 \\
    ConvNeXt-V2-base & 30.9 & 1.1 & 40.8 & 0.42 \\
    ViT-MAE-base & 30.9 & 3.8 & 40.0 & 0.43 \\
    ViT-MSN-large & 31.6 & 5.2 & 40.3 & 0.42 \\
    InternViT-300M & 33.1 & 4.6 & 42.6 & 0.37 \\
    \midrule
    PAD Specialist Baseline  other than LoRA \\
    \midrule
    \htablegrayrow{FLIP-MCL & \bestcell{17.3} & 0.7 & \bestcell{22.9} & \bestcell{0.73}} \\
    FLIP-IT & 19.5 & \bestcell{0.5} & 25.8 & 0.69 \\
    FLIP-V & 21.5 & 1.5 & 28.1 & 0.65 \\
    FSFM-FAS & 29.0 & 4.2 & 37.3 & 0.48 \\
    DeepPixBiS & 33.2 & 3.4 & 43.1 & 0.36 \\
    \bottomrule
  \end{tabular}
  }
\end{table}

\textit{Does lightweight adaptation solve cross-dataset shift?}
While intra-dataset ACER demonstrates strong performance following LoRA adaptation, cross-dataset ACER remains between 20\% and 43\% for all models, irrespective of pretraining regime, parameter count, or intra-dataset rank.
Transfer performance does not correlate with intra-dataset rank. LLaVA-NeXT-7B-ViT achieves the lowest mean cross-dataset ACER (19.8\%), followed by DINOv2-R-giant (20.4\%) and CLIP ViT-B/32 (21.8\%). 
In contrast, InternViT-6B, which is among the strongest intra-dataset backbones, ranks near the bottom (39.8\%).
The choice of training source significantly influences transfer performance. 
For example, CLIP ViT-B/32 adapted on Replay-Attack achieves approximately 0.2\% intra-dataset ACER and approximately 10\% mean cross-dataset ACER, whereas the same model adapted on MSU-MFSD yields approximately 33\% cross-dataset ACER.
\textit{Collectively, these findings indicate that lightweight adaptation does not resolve cross-dataset transfer}. 
Instead, the primary limitation shifts from representation quality to the selection of the backbone and the training dataset for the deployment domain.

\subsection{Pretraining Data Scale}
\label{sec:pretraining-scale}

\begin{figure}[tbp]
  \centering
  \begin{minipage}{0.9\linewidth}\centering
  \subcaptionbox{$\mathrm{ACER}_{\mathrm{ID}}\downarrow$}{
    \includegraphics[width=\linewidth]{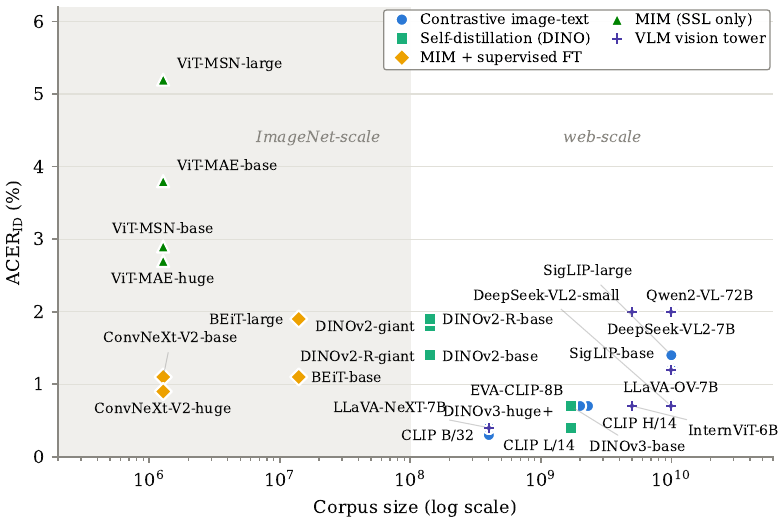}
    }
  \end{minipage} 
  \centering
  \hspace*{-0.02\linewidth}%
  \begin{minipage}{0.96\linewidth}\centering
  \subcaptionbox{$\mathrm{ACER}_{\mathrm{ALL}}\downarrow$}{
    \includegraphics[width=\linewidth]{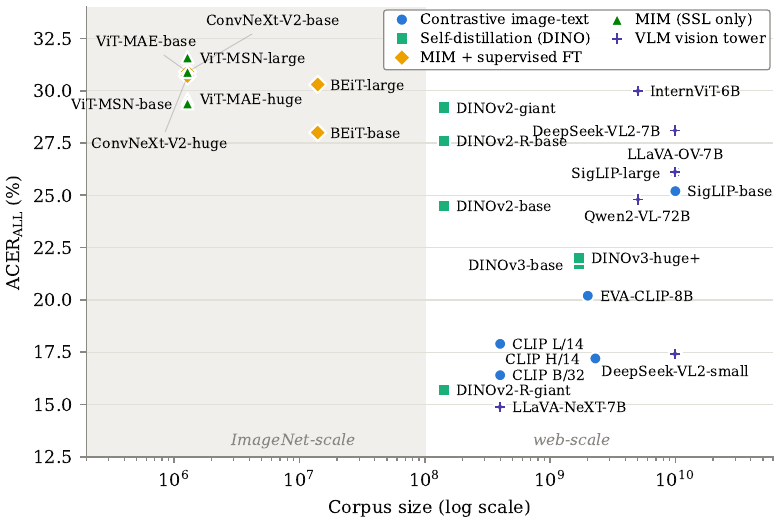}
    }
  \end{minipage}
  \caption{
    Relationship between dataset size and ACER. 
    The mean intra-dataset error (ACER$_{\mathrm{ID}}$, left) and cross-dataset error (ACER$_{\mathrm{ALL}}$, right) are presented. Colors and markers indicate pretraining families (see~\cref{sec:encoderfamily}).  
    The shaded band denotes ImageNet-scale datasets ($\leq 10^8$ images). Intra-dataset performance is insensitive to dataset size, while cross-dataset performance plateaus beyond $10^8$ images. 
    Legend explained in \cref{sec:encoderfamily}.
}
\label{fig:pretraining-scale-id-all}
\end{figure}

\Cref{fig:pretraining-scale-id-all} relates each backbone's pretraining dataset size to its post-LoRA error.
The intra-dataset view (left) is the control for the data-scale argument: after LoRA, every backbone fits its training dataset (0.3--5.2\% ACER), ImageNet-only backbones sit among the best intra-dataset models, and no left--right structure is visible, so pretraining dataset size is uninformative for intra-dataset performance.
The all-pairs aggregate (right) reverses the picture: every backbone pretrained solely on ImageNet-1K/22K ($\leq$14M images, shaded) sits in the bottom third, every backbone in the top ten saw at least 100M web images, and beyond web scale more data stops helping (SigLIP's ${\sim}$10B WebLI images rank behind CLIP's 400M pairs).
Pretraining data scale therefore manifests almost exclusively under domain shift: web-scale coverage is necessary but not sufficient, and curation and objective dominate beyond the threshold.

\subsection{Embedding Structure}

\definecolor{tsneMSU}{HTML}{0072B2}     
\definecolor{tsneCASIA}{HTML}{009E73}   
\definecolor{tsneReplay}{HTML}{984EA3}  
\definecolor{tsneOulu}{HTML}{E69F00}    

\begin{figure}[tbp]
  \centering
  {\footnotesize{(a) Off-the-shelf (frozen backbone)}}\\[0.2em]
  \begin{minipage}{0.32\linewidth}\centering
    \includegraphics[width=\linewidth]{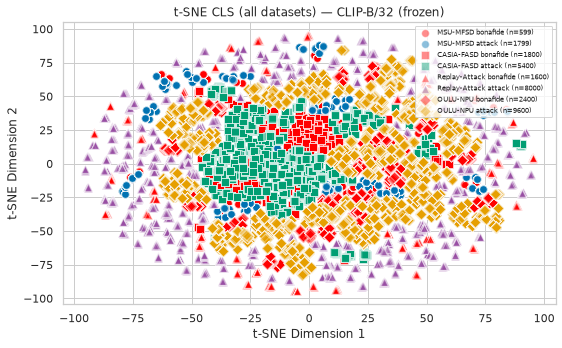}\\[-0.3em]
    {\footnotesize CLIP ViT-B/32}
  \end{minipage}\hfill
  \begin{minipage}{0.32\linewidth}\centering
    \includegraphics[width=\linewidth]{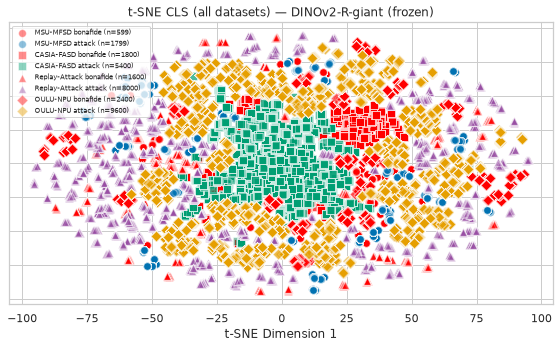}\\[-0.3em]
    {\footnotesize DINOv2-R-giant}
  \end{minipage}\hfill
  \begin{minipage}{0.32\linewidth}\centering
    \includegraphics[width=\linewidth]{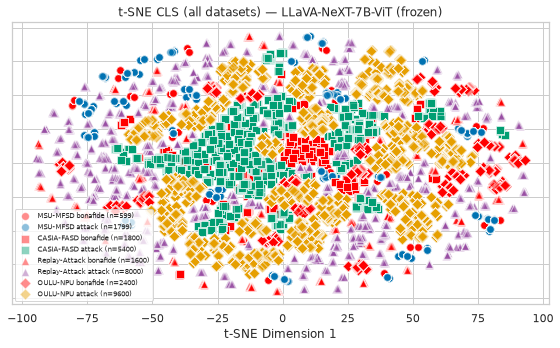}\\[-0.3em]
    {\footnotesize LLaVA-NeXT-7B-ViT}
  \end{minipage}

  \vspace{0.7em}
  {\footnotesize{(b) After LoRA fine-tuning}}\\[0.2em]
  \begin{minipage}{0.32\linewidth}\centering
    \includegraphics[width=\linewidth]{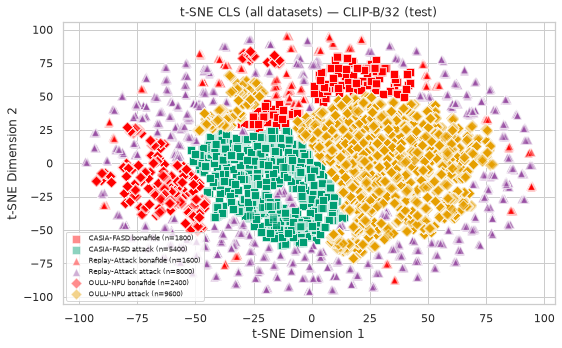}\\[-0.3em]
    {\footnotesize CLIP ViT-B/32}
  \end{minipage}\hfill
  \begin{minipage}{0.32\linewidth}\centering
    \includegraphics[width=\linewidth]{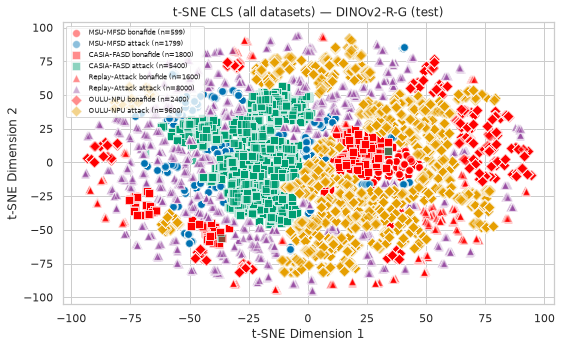}\\[-0.3em]
    {\footnotesize DINOv2-R-giant}
  \end{minipage}\hfill
  \begin{minipage}{0.32\linewidth}\centering
    \includegraphics[width=\linewidth]{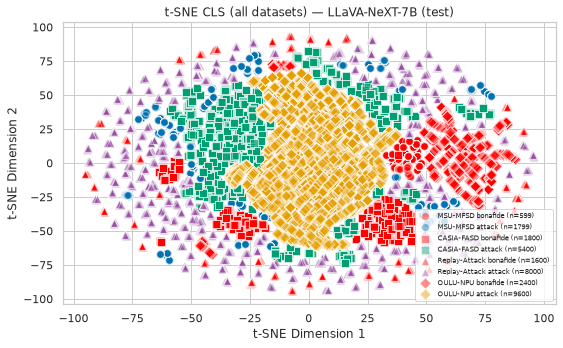}\\[-0.3em]
    {\footnotesize LLaVA-NeXT-7B-ViT}
  \end{minipage}
  \caption{%
    t-SNE of CLS features on the test sets.
    \textbf{(a)} Off-the-shelf frozen backbones (before adaptation);
    \textbf{(b)} the same encoders after LoRA fine-tuning.
    \textcolor{red}{Bonafide} samples are shown in \textcolor{red}{red} across all datasets, while presentation attacks are colored:
    \textcolor{tsneMSU}{MSU-MFSD},
    \textcolor{tsneCASIA}{CASIA-FASD},
    \textcolor{tsneReplay}{Replay-Attack}, and
    \textcolor{tsneOulu}{OULU-NPU}.
    Residual domain structure behind the cross-dataset gap.
  }
  \label{fig:tsne-top}
\end{figure}


\Cref{fig:tsne-top} presents the adapted embeddings of the three most effective backbone models (CLIP ViT-B/32, DINOv2-R-giant, LLaVA-NeXT-7B-ViT) across the test sets.
Although the attention projections are adapted, the feature geometry is still primarily structured by acquisition domain rather than by presentation-attack class. Samples cluster according to their source dataset rather than the bonafide or attack label.
This representational structure accounts for the persistent cross-dataset performance degradation observed in \cref{tab:lora-cross-dataset-summary}. While the low-rank update enhances separation within individual datasets, it does not create a unified spoof-versus-live representation across datasets. As a result, thresholds calibrated in one domain are unlikely to generalize effectively to another.


\subsection{Discussion}


\textbf{Why does adaptation saturate intra-dataset but not cross-dataset?}
Zero-shot VLM prompting does not achieve viable performance (\cref{tab:zeroshot-intra-dataset-acer}), while LoRA adaptation enables strong discrimination within individual acquisition pipelines (\cref{tab:lora-intra-dataset-acer}).
LoRA primarily enhances intra-dataset discrimination, whereas cross-dataset invariance requires additional mechanisms.
Nevertheless, LoRA adaptation does not eliminate acquisition-specific cues that dominate the learned representations.
The resulting embeddings are still organized by the source dataset rather than by the spoof label (\cref{fig:tsne-top}), leading to thresholds calibrated on one dataset that fail under distribution shift.
As a result, the cross-dataset ACER remains between 20 and 43 percent across all model backbones (\cref{tab:lora-cross-dataset-summary}), despite very low intra-dataset error rates. 
For example, InternViT-6B achieves 0.7 percent intra-dataset ACER but 39.8 percent cross-dataset ACER.


\textbf{Why do scale and pretraining volume not rescue transfer?}
Cross-dataset transfer is determined more by the pretraining objective than by model size or pretraining scale. Contrastive vision-language encoders (CLIP, EVA-CLIP) and the DINO family, particularly the register-token variant DINOv2-R-giant, consistently outperform similarly sized ViT-MAE, ViT-MSN, and ConvNeXt-V2 models (\cref{tab:lora-cross-dataset-summary}). These findings align with evidence that PAD benefits from encoders preserving stable, localizable spatial structure~\cite{darcet2024registers,feng2026benchmarking}. Increasing model size provides little benefit: the 0.09B CLIP ViT-B/32 surpasses every multi-billion-parameter backbone, and smaller variants often transfer better within the same family. 
Likewise, increasing pretraining data beyond web scale yields no consistent improvement (\cref{sec:pretraining-scale}, \cref{fig:pretraining-scale-id-all}). This trend is also evident among VLM vision towers, where transfer performance tracks the degree of proximity to the original contrastive initialization, from the lightly tuned CLIP tower in LLaVA-NeXT (19.8\%) to the heavily adapted Qwen towers (32.4--38.6\%).




\section{Conclusion}
\label{sec:conclusion}

We presented a unified benchmark for face PAD, combining LoRA adaptation across 32 vision encoders with zero-shot evaluation of nine complete VLMs under a common protocol on four datasets and against specialist baselines DeepPixBiS and FSFM-FAS.

Our results highlight several key observations. 
LoRA adaptation substantially improves intra-dataset performance. 
Most adapted backbones achieve intra-dataset ACER below 2\% (\cref{tab:lora-intra-dataset-acer}), whereas zero-shot prompting of the same models remains near chance (\cref{tab:zeroshot-intra-dataset-acer}).
Cross-dataset transfer remains challenging after adaptation: ACER ranges from 20--43\% and is strongly influenced by the choice of training dataset (\cref{tab:lora-cross-dataset-summary}).
Moreover, increasing model scale alone does not consistently improve transferability, as seen with InternViT-6B. 
The vision encoder of DeepSeek-VL2 contains SigLIP that is originally trained on ImageNet-scale, but in combination in multi-stage training in a whole multimodal model, its performance has increased. 
Most importantly, the rather small CLIP ViT-B/32 and the much larger extracted CLIP-vision tower from the MLLM LLaVA-NeXT-7B achieve the strongest detection accuracy, whereas CLIP ViT-B/32 outperforms in overall compute~\cref{fig:compute-efficiency}.

These findings raise concerns regarding the effectiveness of previous approaches~\cite{gonzalez2025foundation,wang2025fsfm} that utilize additional data, such as leave-one-out (LOO) or auxiliary datasets, to enhance generalization.
Focusing on other post-training methods, including additional fine-tuning with domain-invariant losses or alignment with target-domain data, may mitigate cross-dataset issues. \\

\textbf{Limitation.} 
Our study is limited to the MCIO benchmark, which primarily covers print and replay attacks and does not include deepfake-based attacks. Cross-dataset generalization is evaluated under a single-source protocol, and lightweight adaptation is restricted to LoRA rather than a broader comparison of parameter-efficient fine-tuning methods. Moreover, zero-shot evaluation uses a single prompt formulation, leaving prompt sensitivity unexplored. 

\section*{Acknowledgment}
This research was funded by the European Union project CarMen (Grant Agreement No. 101168325).






\FloatBarrier
\bibliographystyle{splncs04}

\end{document}